\documentclass{article}

\PassOptionsToPackage{numbers, sort&compress}{natbib}

\usepackage[preprint]{neurips_2026}

\usepackage[utf8]{inputenc} %
\usepackage[T1]{fontenc}    %
\usepackage{url}            %
\usepackage{microtype}
\usepackage{graphicx}
\usepackage{subcaption}
\usepackage{placeins}
\usepackage{booktabs} %
\usepackage{amsfonts}
\usepackage{nicefrac}
\usepackage{xcolor}
\usepackage{multirow}
\usepackage{wrapfig}
\usepackage{xspace}   %

\usepackage{hyperref}
\hypersetup{hidelinks}
\usepackage{tikz}
\usetikzlibrary{shapes,arrows,positioning,fit,backgrounds,calc,decorations.pathreplacing,matrix}
\usepackage{pifont}
\usepackage{algorithm}
\usepackage{algorithmic}
\usepackage{enumitem}
\setlist[itemize]{leftmargin=*}
\setlist[enumerate]{leftmargin=*}
\usepackage{tcolorbox}

\usepackage{amsmath}
\usepackage{amssymb}
\usepackage{mathtools}
\usepackage{amsthm}
\usepackage[capitalize,noabbrev]{cleveref}

\newcommand{\oursystem}{\mbox{\rmfamily\upshape HypoEvolve}\xspace}
\newcommand{\depmap}{\mbox{\fontfamily{lmss}\upshape\selectfont DepMap}\xspace}
\newcommand{\drugname}[1]{{\rmfamily\textsc{\MakeLowercase{#1}}}}

\theoremstyle{plain}

\theoremstyle{definition}

\theoremstyle{remark}

\title{
\oursystem: Genetic Algorithms Enable\\
Multi-Agent LLMs to Discover Scientific Hypotheses
}

\author{%
  \bfseries
  Jieyuan Liu\textsuperscript{1*} \quad
  Mengzhou Hu\textsuperscript{1} \quad
  Jefferson Chen\textsuperscript{1} \quad
  JungHo Kong\textsuperscript{1}
  \\[2pt]
  \bfseries
  Pratibha Jagannatha\textsuperscript{1} \quad
  Yiming Gao\textsuperscript{2} \quad
  Dexter Pratt\textsuperscript{1} \quad
  Hsin-Yuan Lee\textsuperscript{1}
  \\[2pt]
  \bfseries
  Zhiting Hu\textsuperscript{1} \quad
  Trey Ideker\textsuperscript{1} \quad
  Wei Wang\textsuperscript{1} \quad
  Eric P. Xing\textsuperscript{3,4} \quad
  Zhen Wang\textsuperscript{1*}
  \\[6pt]
  {\normalfont\small
    \textsuperscript{1}University of California San Diego
    \quad
    \textsuperscript{2}Texas A\&M University
  }
  \\[-1pt]
  {\normalfont\small
    \textsuperscript{3}Carnegie Mellon University
    \quad
    \textsuperscript{4}Mohamed bin Zayed University of Artificial Intelligence
  }
}

\begin{document}

\raggedbottom

\maketitle

\begingroup
\renewcommand{\thefootnote}{*}
\footnotetext[1]{Correspondence: \href{mailto:jil029@ucsd.edu}{jil029@ucsd.edu}, \href{mailto:zhenwang.work@gmail.com}{zhenwang.work@gmail.com}}
\endgroup

\begin{abstract}

Scientific agents increasingly contribute to hypothesis discovery by synthesizing evidence, assessing proposals, and developing new explanations.
Recent systems bring scientific agents and evolutionary search together to develop hypotheses through cycles of critique, comparison, and revision.
However, how different forms of agent collaboration affect hypothesis quality remains an open question.
Answering this question requires separating the effects of agents' scientific capabilities from those of their collaboration.
A suitable framework must therefore preserve the agents' scientific roles and support different rules for combining, revising, and retaining hypotheses.
Building on this perspective, we introduce \oursystem, which makes collaboration explicit through successive updates to a hypothesis population.
Specifically, we propose to use a generational genetic algorithm to coordinate specialized large language model (LLM) agents that integrate mechanistic arguments, reconsider assumptions, and assess evidence and testability.
Each generation specifies how scientific judgments and new proposals reshape the population, which makes the effects of collaboration on hypothesis quality directly testable.
Moreover, we design our evaluation around scientifically meaningful hypotheses that explain how a proposed intervention could work.
Drug repurposing connects these explanations to target-level biological claims that can be assessed against external evidence.
Specifically, we adapt \depmap and Open Targets into complementary external measures grounded in experimental, genetic, and clinical evidence.
The evaluation spans 34 cancer types, with \oursystem achieving the highest scores against six baselines on both measures.
\depmap selectivity reaches 0.171, compared with 0.115 for the strongest baseline.
Gains over single-pass generation also generalize to held-out cancer types.
\oursystem advances a vision of autonomous science in which AI research teams achieve a capacity for discovery beyond that of individual models.

\end{abstract}

\section{Introduction}

Large language models (LLMs) are enabling scientific agents to formulate hypotheses that connect existing evidence to new research directions \citep{wang2024scimon,yang2025moosechem,gottweis2025coscientist}. They can synthesize findings across studies into explicit scientific claims and supporting rationales that connect proposed relationships to the available evidence \citep{baek2025researchagent,ghafarollahi2025sciagents}. These capabilities open a path to systems that develop scientific ideas through repeated examination of hypotheses and their supporting evidence \citep{gottweis2025coscientist,ghareeb2025robin,gao2025scpilot}.

Recent progress in automated discovery spans scientific-agent workflows and evolutionary search. One line of research develops agents that ground proposals in the literature and refine them through critical feedback \citep{wang2024scimon,baek2025researchagent,ghafarollahi2025sciagents}. Another uses evolutionary search to develop LLM-generated programs, equations, and molecules, with evaluation and selection guiding subsequent exploration \citep{romeraparedes2024funsearch,novikov2025alphaevolve,wang2026evodiverse}. Recent systems bring these directions together for scientific hypotheses through tournament-based evolution and hierarchical refinement \citep{gottweis2025coscientist,yang2025moosechem2}. Yet it remains unclear how the design of agent collaboration affects the hypotheses a research team develops. A system's performance reflects both the agents' scientific capabilities and the decisions that direct their work. Isolating the contribution of collaboration would provide a basis for designing teams with scientific capabilities \mbox{beyond those of their individual members}.

\begin{samepage}
Controlled comparisons require scientific roles and search decisions to be specified separately \citep{hao2023rap,hao2024reasoners}. Scientific-agent systems assign generation, critique, and synthesis to specialized roles \citep{ghafarollahi2025sciagents,gottweis2025coscientist}. Evolutionary algorithms provide explicit rules for selecting and varying candidate solutions \citep{eiben2015evolutionary,novikov2025alphaevolve}. Our formulation makes these rules govern how agents develop a population of hypotheses. Each population update determines which proposals agents receive and which outputs enter the next round. We can then vary the search rules with scientific roles, prompts, and evaluation criteria held fixed, making collaboration an experimental variable and hypothesis quality the outcome.

\end{samepage}

\begin{figure}[!t]
    \centering
    \includegraphics[trim=40bp 0bp 51bp 0bp,clip,width=\linewidth]{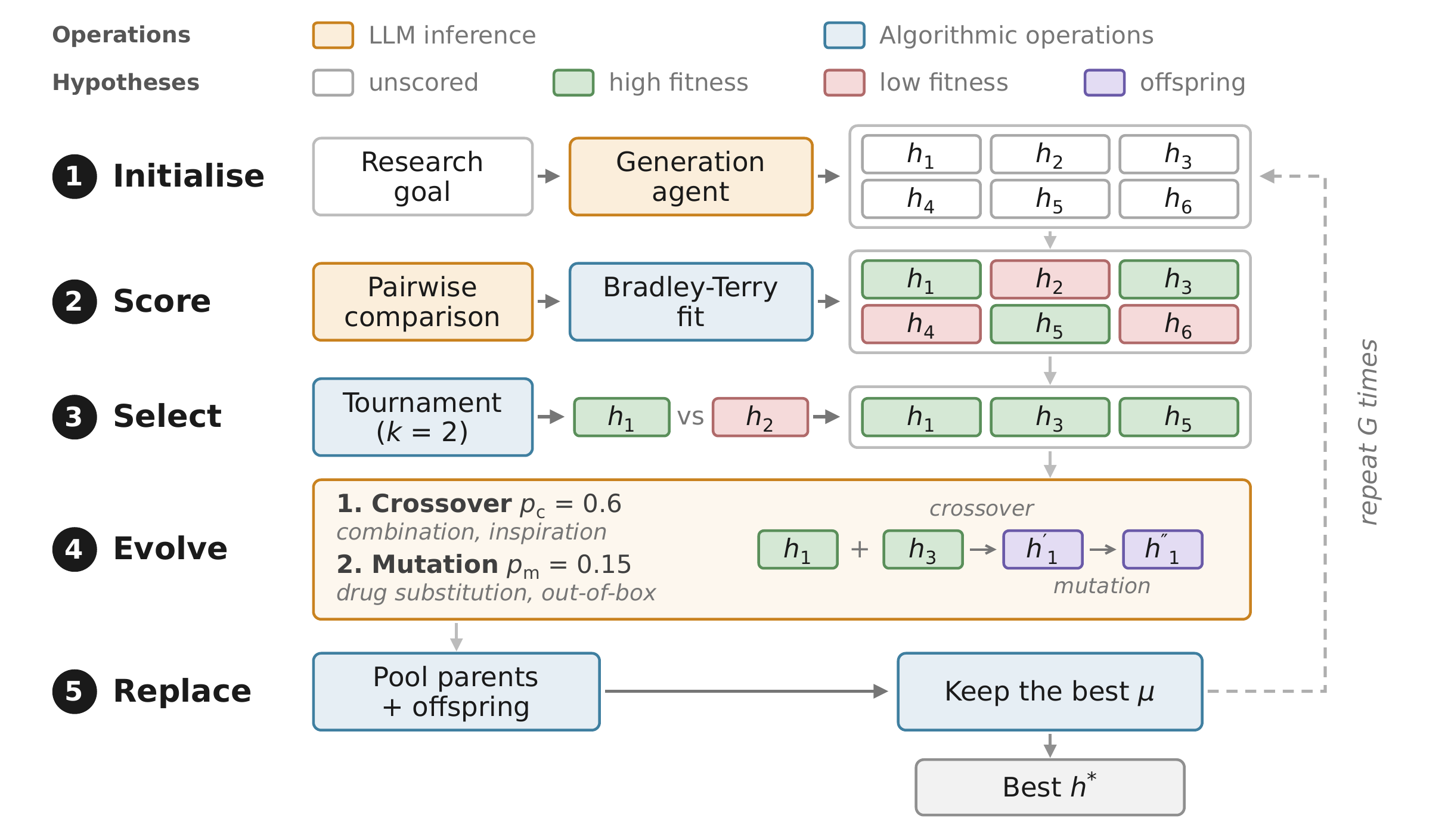}
    \caption{\textbf{Overview of \oursystem.} The genetic algorithm connects agents' scientific judgments to the hypotheses developed in the next generation, coordinating evaluation, semantic variation, and population replacement. Orange and blue nodes denote LLM calls and algorithmic operations.}
    \label{fig:motivation}
\end{figure}

To realize this formulation, we propose \oursystem (Figure~\ref{fig:motivation}), a generational genetic framework in which specialized LLM agents provide both scientific variation and comparative fitness. We use pairwise judgments of evidence and testability to direct exploration toward promising hypotheses. To develop substantive scientific alternatives, we formulate crossover and mutation as reasoning over claims and rationales. Agents can combine mechanistic arguments across hypotheses or reconsider the assumptions behind an explanation. We evaluate parents and offspring together and retain a fixed-size population, so new proposals compete directly with the ideas they build on. This replacement rule connects comparative judgment to the direction of subsequent search. We record parentage and operator choices to make each hypothesis's development inspectable across generations. The resulting framework makes the coordination of scientific agents explicit and supports controlled changes to the search without redefining their scientific roles.

We evaluate hypothesis discovery through the biological implications of proposed scientific explanations. Because prospective experiments are costly \citep{gottweis2025coscientist,wang2026firebench}, we construct a drug repurposing evaluation that connects candidate interventions and mechanistic rationales to external evidence \citep{ashburn2004repurposing,pushpakom2019repurposing,yue2020bionev}. Each rationale implies that the drug's targets are relevant to the specified cancer, providing a concrete biological claim for assessment. Across 34 cancer types \citep{weinstein2013tcga}, we assess this claim with \depmap selectivity \citep{tsherniak2017depmap,meyers2017ceres} and Open Targets association \citep{ochoa2023opentargets}, reserving both measures for use after the search. Under a shared task and retrieval protocol, \oursystem achieves the highest mean scores among six baselines on both measures. \depmap selectivity reaches 0.171 and Open Targets association reaches 0.426, compared with 0.115 and 0.329 for Tree of Thoughts, the strongest baseline \citep{yao2023tree}. The advantage over single-pass generation generalizes to held-out cancer types. With scientific operations and hypothesis count held fixed, fitness-guided parent selection improves the population's mean and minimum scores. This connection between collaboration design and hypothesis quality offers a foundation for building autonomous AI research teams.

\section{Related Work}

\noindent\textbf{Scientific Hypothesis Discovery.} Literature-based discovery generates hypotheses by connecting findings across scientific studies \citep{swanson1986fishoil,spangler2014automated,sybrandt2017moliere}. LLM systems now make hypotheses explicit natural-language artifacts that can be generated, evaluated, and revised. HypoGeniC iteratively updates hypotheses from labeled examples \citep{zhou2024hypothesis}, SciMON optimizes literature-grounded scientific directions for novelty \citep{wang2024scimon}, and ResearchAgent uses reviewing agents to refine research proposals \citep{baek2025researchagent}. A large-scale expert study further shows that novelty, feasibility, and self-evaluation capture different dimensions of research-idea quality \citep{si2025ideas}. At the level of complete research workflows, the AI Scientist automates idea generation, experimentation, analysis, and manuscript writing \citep{lu2024aiscientist}; its template-free variant uses agentic tree search to develop experimental implementations \citep{yamada2025aiscientistv2}. Agent Laboratory carries a researcher-provided idea through literature review, experimentation, and report generation \citep{schmidgall2025agentlaboratory}. \oursystem targets the upstream problem of developing scientific hypotheses, using generational search to refine claims whose biological implications \mbox{are assessed against external evidence}.

\noindent\textbf{Multi-Agent Systems for Scientific Hypothesis Discovery.} Multi-agent scientific systems distribute generation, criticism, synthesis, and prioritization across specialized roles \citep{ghafarollahi2025sciagents,wang2026cellmaster,hu2026tritondft}. MOOSE-Chem retrieves scientific inspirations and composes chemistry hypotheses \citep{yang2025moosechem}, SciAgents combines ontological knowledge graphs with collaborating agents for materials research \citep{ghafarollahi2025sciagents}, and multi-agent LLMs have generated drug-combination hypotheses \citep{qi2025hypothesis}. Robin connects hypothesis formation to experimental feedback \citep{ghareeb2025robin}. Co-Scientist is the closest multi-agent reference point. It uses generation, debate, ranking, and evolution agents, and ranks hypotheses through an Elo-based tournament within an expanding pool \citep{gottweis2025coscientist}. The Hypothesis Evolution Protocol separately records hypothesis generation, testing, evidence, and belief updates in an auditable registry \citep{takahara2026hep}. \oursystem defines collaboration through a fixed-size generational genetic search, with explicit parent selection, controlled semantic variation, joint parent-offspring replacement, and recorded lineages.

\noindent\textbf{Evolutionary Search over Language Artifacts.} Evolutionary methods increasingly treat language-model artifacts as members of a population. EvoPrompt and Promptbreeder evolve prompts \citep{guo2023evoprompt,fernando2024promptbreeder}, while Evolution through Large Models and FunSearch evolve executable programs \citep{lehman2022elm,romeraparedes2024funsearch}. Quality-Diversity through AI Feedback extends population search to diverse text \citep{bradley2024qdaif}. Language Model Crossover provides a general crossover operator for text-representable artifacts, including sentences, equations, prompts, and code \citep{meyerson2024crossover}. Related approaches evolve agent teams \citep{yuan2024evoagent} or train agents jointly through co-evolution \citep{chen2025multiagentevolve}. EvoDiverse brings population-based exploration to scientific-hypothesis search \citep{wang2026evodiverse}. It uses multiple temperature-controlled populations and swap rules to optimize quality and diversity under a fixed validation budget, with experiments over molecules, equations, and algorithms scored by domain-specific automated oracles that drive selection. \oursystem evolves structured scientific claims under agent-derived fitness, separating generational genetic search from subsequent assessment against external biological evidence.

\section{Method}

\noindent\textbf{Problem Formulation.} Given a natural-language research goal $g$, we seek hypotheses that address the goal with scientifically grounded explanations and potentially new insights. Each candidate $h$ is a structured document containing a title, summary, hypothesis statement, and supporting rationale. We formulate discovery as a finite population search with $\mu$ retained candidates, $\lambda$ offspring per generation, and a horizon of $G$ generations. Let $P_t$ denote the population at generation $t$ and $f_t$ the fitness inferred from task-specific comparisons in that generation. The search returns the highest-fitness hypothesis in the final population,
\begin{equation}
    h^* = \arg\max_{h \in P_G} f_G(h).
    \label{eq:search-output}
\end{equation}
Fitness summarizes the agents' assessments under the specified scientific criteria and directs parent selection and population replacement. External biological evidence is applied only after search to assess the resulting drug repurposing hypotheses (Section~\ref{sec:setup}).

\begin{samepage}
\noindent\textbf{Algorithm Overview.} \oursystem separates reasoning over scientific content from the population update that coordinates it. Agents supply hypothesis generation, semantic variation, and comparative fitness; the genetic algorithm specifies how these outputs change the population \citep{holland1992adaptation,eiben2015evolutionary}. A generation agent initializes $P_0$ from retrieved literature. At generation $t$, selected parents produce $\lambda$ offspring $O_t$ through LLM-based crossover and mutation. A pairwise scorer evaluates parents and offspring together, and a deterministic supervisor retains the top $\mu$ candidates,
\begin{equation}
    Q_t = P_{t-1} \uplus O_t, \qquad
    f_t = \operatorname{Score}(Q_t), \qquad
    P_t = \operatorname{Top}_{\mu}(Q_t;f_t).
    \label{eq:population-update}
\end{equation}
\end{samepage}
Here $\uplus$ pools candidate records, preserving distinct identities even when their text is unchanged. Lineage records support traceability, while comparative fitness guides selection. Search decisions alter the hypotheses supplied to the comparison and evolution agents while their role definitions, prompts, and scientific criteria remain fixed. The parent-selection study in Section~\ref{sec:ablations} uses this separation to change a search rule while preserving the scientific operators and hypothesis count. Figure~\ref{fig:motivation} depicts the agent calls, and Appendix~\ref{app:algorithm} formalizes the full search in Algorithm~\ref{alg:hypoevolve}.

\subsection{\texorpdfstring{LLM Agents as Semantic Search Operators}{LLM Agents as Semantic Search Operators}}

\noindent Three specialized agents implement generation, comparison, and evolution. They operate on the claims and rationales within each hypothesis, allowing genetic operations to act on scientific content. Appendix~\ref{app:prompts} provides the prompts for each role in our drug-repurposing instantiation.

\noindent\textbf{Literature-Grounded Initialization.} The generation agent derives literature queries from $g$, retrieves relevant papers, and synthesizes their findings. It uses this evidence to propose $\mu$ hypotheses spanning different mechanisms, pathways, and interventions. Each proposal follows the same structured format, so later agents receive both a scientific claim and the rationale supporting it.

\noindent\textbf{Comparative Scientific Judgment.} The pairwise scorer compares two hypotheses under task-specific criteria and selects the stronger candidate or declares a tie. Pairwise judgments offer a practical basis for ranking open-ended language outputs \citep{zheng2023judge,liusie2024pairwise,yin2026decentralized}. For drug repurposing, the scorer considers specificity to the named cancer, evidence implicating the proposed target, and whether the hypothesis makes a concrete, falsifiable prediction. These criteria direct attention to cancer-specific dependencies and the scientific argument for each drug repurposing hypothesis. \depmap and Open Targets data are reserved for external assessment and do not enter the scorer.

\noindent\textbf{Semantic Crossover.} The evolution agent develops offspring from two selected parents using language-model crossover \citep{meyerson2024crossover}. The \emph{combination} operator integrates compatible mechanisms or evidence from both parents into a coherent explanation. The \emph{inspiration} operator uses their ideas as starting points for a different explanation aligned with the research goal.

\noindent\textbf{Semantic Mutation.} Mutation develops a single hypothesis by revising its proposed intervention or reconsidering its explanation. In the drug-repurposing instantiation, \emph{drug substitution} changes the proposed compound within the allowed vocabulary while retaining the mechanistic argument. The \emph{out-of-box} operator revisits the hypothesis's assumptions and explores alternative explanations.

\subsection{\texorpdfstring{Generational Search with Comparative Fitness}{Generational Search with Comparative Fitness}}
\label{sec:genetic-search}

\noindent The supervisor turns these agent operations into an explicit generational search. It determines which hypotheses reproduce, which variation operators act on them, and which candidates remain in the population. The same procedure repeats at every generation, with parent and operator records tracing the origin of each offspring.

\noindent\textbf{Population-Level Fitness.} Population fitness aggregates pairwise scientific judgments into a ranking. The scorer compares every unordered pair in $P_0$ at initialization and in the parent-offspring pool $Q_t$ at each subsequent generation. A Bradley-Terry model \citep{bradley1952rank} converts the comparison outcomes into positive latent strengths $\pi_{t,h}$ and the resulting search fitness,
\begin{equation}
    P(h_i \succ h_j) = \frac{\pi_{t,h_i}}{\pi_{t,h_i} + \pi_{t,h_j}},
    \qquad f_t(h) = a\log\pi_{t,h} + b_t, \quad a>0.
    \label{eq:comparative-fitness}
\end{equation}
The initial fit sets the population mean to 50 and the spread to 60 points, fixing the multiplier $a$ for the run. Later fits retain this multiplier and adjust only the offset $b_t$ to align with the previous scores of surviving candidates. This anchoring supplies a common within-run reference for fitness trajectories; selection uses the ordering within each comparison pool.

\noindent\textbf{Fitness-Guided Reproduction.} Each parent selection samples two distinct candidates uniformly from $P_{t-1}$ and chooses the one with higher fitness \citep{miller1995genetic}. Crossover draws two parents through separate tournaments, resampling the second if it matches the first. With six strictly ranked hypotheses, the strongest wins a third of tournaments and the fifth-ranked wins one in fifteen. The tournament therefore favors stronger candidates while allowing every member except the weakest to reproduce.

For each offspring, crossover is applied with probability $p_c=0.6$. Mutation then acts on the result with probability $p_m=0.15$, or with probability 1 if crossover was skipped. Each operation selects uniformly between its two variants. An empty operator return triggers an unchanged parent copy, preserving the offspring count. Every offspring receives a separate record with its parentage and operator provenance, including these fallback copies.

\noindent\textbf{Joint Parent-Offspring Replacement.} The supervisor scores the $\mu$ parents and $\lambda$ offspring together and retains the top $\mu$, implementing $(\mu+\lambda)$ truncation \citep{beyer2002evolution}. Parents remain eligible alongside their descendants, so a new proposal enters the retained population by ranking among the strongest candidates in the combined pool. This joint comparison links semantic variation to population change and supplies the parents for the next generation of hypothesis development.

\section{Experiments}
\label{sec:experiments}

\begin{figure}[!t]
    \centering
    \includegraphics[width=\textwidth]{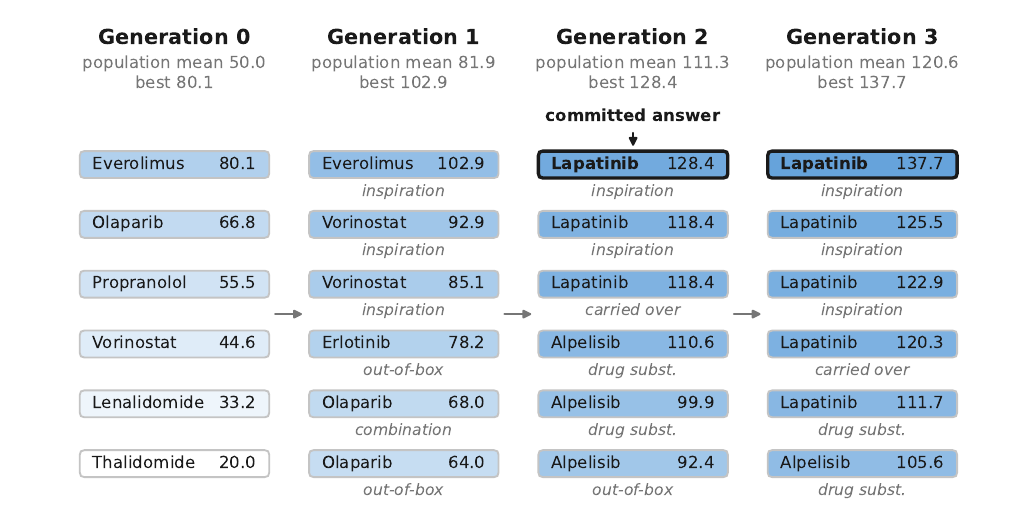}
    \caption{\textbf{A hypothesis population across generations.} A \drugname{lapatinib} hypothesis introduced by inspiration leads the population in generations 2 and 3, illustrating how new proposals redirect the search. Columns show six retained hypotheses in a uterine corpus endometrial carcinoma run; numbers are fitness scores and labels identify \mbox{generating operators or unchanged carryovers}.}
    \label{fig:methodology}
\end{figure}

\subsection{Experimental Setup}
\label{sec:setup}

\noindent\textbf{Task Definition.} We assess whether hypothesis development identifies interventions supported by independent biological evidence. Drug repurposing makes this question concrete by asking whether an existing compound could act on a disease-specific vulnerability \citep{ashburn2004repurposing,pushpakom2019repurposing}. Each hypothesis proposes a drug candidate and explains how its targets or pathways could affect the specified cancer. This explanation entails an assessable biological implication, namely that the implicated targets are relevant to that cancer. We test this implication through CRISPR perturbations and curated target-disease associations. These scores measure biological support for the proposed drug repurposing opportunity; prospective experiments are needed to establish the full mechanism and therapeutic benefit.

\noindent\textbf{Dataset.} Our evaluation spans 34 cancer types, covering the 33 represented in The Cancer Genome Atlas (TCGA) \citep{weinstein2013tcga} and chronic myelogenous leukemia. Paired comparisons use the 29 cancer types for which every method produced an answer. Three types, kidney chromophobe (KICH), pheochromocytoma and paraganglioma (PCPG), and thymoma (THYM), have no matching \depmap cell lines, leaving 26 types for \depmap selectivity and 29 for Open Targets association.

\noindent\textbf{Held-Out Protocol.} Four cancer types informed protocol development, namely acute myeloid leukemia, breast invasive carcinoma, pancreatic adenocarcinoma, and skin cutaneous melanoma. Three more appeared in an interim inspection of the frozen batch, namely adrenocortical carcinoma, bladder urothelial carcinoma, and brain lower grade glioma. We exclude all seven from the held-out analysis. The remaining 27 types were evaluated with no further configuration changes, providing a test beyond the cancer contexts used during protocol development and inspection.

\noindent\textbf{Evaluation Metrics.} \depmap CRISPR screens measure how strongly cancer cell lines depend on individual genes for survival \citep{tsherniak2017depmap,meyers2017ceres}. Raw target dependency can reward genes that are essential across many cancers. A constant \drugname{thalidomide} answer ranks first, with ties, in 30 of the 31 cancer types with matched cell lines under this score. We therefore measure selectivity relative to each target's pan-cancer dependency. \drugname{Thalidomide} in acute myeloid leukemia falls from 1.0000 to $+0.0040$, while \drugname{imatinib} in chronic myelogenous leukemia retains $+0.9674$ and \drugname{vemurafenib} in melanoma $+0.9378$. The two external metrics are:
\begin{itemize}
    \item \textbf{\depmap selectivity}: For each drug, we subtract each target's pan-cancer median dependency from its median in the matched cancer and take the maximum across annotated targets.
    \item \textbf{Open Targets association} \citep{ochoa2023opentargets}: The association score between the drug's annotated targets and the matched cancer, providing evidence independent of CRISPR screens.
\end{itemize}

\noindent\textbf{Evaluation Protocol.} Each run selects one drug repurposing hypothesis before external scoring. For \oursystem, this is the highest-fitness hypothesis in the final population; every baseline likewise returns one hypothesis and its proposed drug. Scores are averaged within each cancer type before paired comparisons, giving cancer types equal weight. No method is evaluated by taking an externally selected maximum over its candidate pool. \depmap and Open Targets scores are computed after candidate selection and never enter search fitness. All methods use the same curated vocabulary of 61 drugs with annotated targets covered by \depmap (Appendix~\ref{app:prompts}).

\noindent\textbf{Baselines.} Six task-matched baselines cover independent generation, sampling, reranking, agentic revision, and tree search. They share the base model, drug vocabulary, retrieval protocol, and single-hypothesis output format. Table~\ref{tab:cost} reports computational costs; Appendix~\ref{app:comparison-details} details the scoring protocol and tests alternative retrieval and answer-selection settings.

\noindent\textbf{(1)~Single-pass generation} produces one hypothesis per run.
\textbf{(2)~Self-consistency} tests agreement across 40 independent samples \citep{wang2023selfconsistency}.
\textbf{(3)~Static reranking} selects from a fixed pool of 15 candidates without iterative refinement \citep{snell2025scaling}.
\textbf{(4)~Multi-agent debate} refines hypotheses through critique and revision \citep{du2023multiagent}.
\textbf{(5)~Co-scientist scaffold} uses generation, ranking, and meta-review in an expanding pool \citep{gottweis2025coscientist}. Feedback guides subsequent proposals, without crossover, mutation, or population replacement.
\textbf{(6)~Tree of Thoughts} uses beam search to develop and select hypotheses \citep{yao2023tree}.

\noindent\textbf{Implementation Details.} All agents use gpt-5.4-mini, with structured prompts and Tavily retrieval of literature relevant to each research goal. We use a population of $\mu=6$, produce $\lambda=6$ offspring per generation, and run for $G=3$ generations, with crossover probability $p_c=0.6$ and mutation probability $p_m=0.15$. We run \oursystem two or three times per cancer type, yielding 94 runs, and average six independent draws for single-pass generation. The sensitivity analysis also tests a population of ten and a horizon of five generations. We fixed the configuration independently of the sensitivity tests and ablations in Section~\ref{sec:ablations}.

\subsection{Main Results}
\label{sec:results}

\begin{figure}[!t]
    \centering
    \includegraphics[width=\columnwidth]{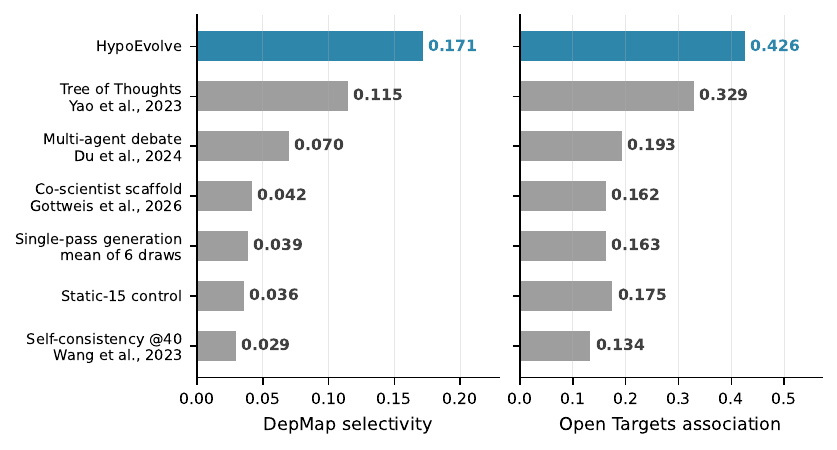}
    \caption{\textbf{Comparison with six hypothesis-discovery baselines.} \oursystem achieves the highest mean on both biological metrics, with Tree of Thoughts the strongest baseline. Means cover 26 cancer types for \depmap selectivity and 29 for Open Targets. Each run contributes one hypothesis selected by the method before external scoring.}
    \label{fig:depmap_summary}
\end{figure}

\noindent\textbf{\oursystem achieves the highest mean scores.} On the 26 cancer types covered by every method and \depmap (Figure~\ref{fig:depmap_summary}), selectivity reaches 0.171 against 0.039 for single-pass generation, a paired margin of $+0.133$ with 19 wins and 7 losses. On the 29 types shared by all methods and Open Targets, association reaches 0.426 against 0.163, a margin of $+0.263$ with 26 wins and 3 losses. Tree of Thoughts is the strongest baseline on both metrics, scoring 0.115 and 0.329, with a \depmap margin of $+0.057$ in favor of \oursystem. Rankings differ only among the closely grouped co-scientist scaffold, single-pass generation, and static reranking, whose scores lie within 0.007 on \depmap and 0.013 on Open Targets. The ordering otherwise agrees across the two sources of biological evidence.

\noindent\textbf{The gains generalize to held-out cancer types.} Across the 27 held-out cancer types, \oursystem exceeds single-pass generation by $+0.280$ on Open Targets and by $+0.111$ on the 24 types with matching \depmap cell lines. All seven development or interim-inspection types are excluded. These margins assess the same frozen configuration beyond the cancer contexts used to develop and inspect the protocol, supporting transfer within the evaluated application domain. Appendix~\ref{app:statistics} reports the paired comparisons and held-out statistics.

\noindent\textbf{The advantage is strongest on genetic evidence.} We separate Open Targets evidence sources to examine whether the advantage is concentrated in channels that directly document drug-disease pairs. Across the 34 cancer types where both methods produced an answer, \oursystem exceeds single-pass generation by $+0.247$ on known drugs and clinical trials, with 26 wins and 5 losses; by $+0.313$ on literature, with 30 wins and 4 losses; and by $+0.334$ on genetic association, with 25 wins and 4 losses. The strongest margin in genetics extends the advantage beyond directly documented drug-disease evidence. Prior exposure may still contribute to these gains.

\subsection{Hypothesis Evolution}
\label{sec:dynamics}

\noindent\textbf{Evolved hypotheses better match drugs to cancer types.} We test whether evolution produces drug candidates whose biological evidence is more specific to the proposed cancer. For each external metric, we compare a drug's score in that cancer with its mean score across other cancers. The resulting residual measures how well the proposed cancer matches the drug's biological evidence. Table~\ref{tab:specificity} reports the full trajectory over 31 cancer types on \depmap and 34 on Open Targets. Both metrics show their largest increase after the first round. The \depmap residual increases through generation 3, while the Open Targets residual peaks at generation 2. From initialization to the final generation, matching improves in 21 of 31 cancer types on \depmap and 25 of 34 on Open Targets. Appendix~\ref{app:specificity} gives the statistical comparisons and controls for drugs that score highly across cancers.

\begin{table}[!t]
    \caption{\textbf{Drug-cancer matching across generations.} Mean residuals compare a proposed drug's score in the matched cancer with its average across other cancer types. Both metrics increase from the initial to the final generation, indicating better drug-cancer matching after accounting for drugs that score highly across many cancers.}
    \label{tab:specificity}
    \centering
    \begingroup\color{black}
    \begin{tabular}{crr}
    \toprule
    Generation & \depmap selectivity ($n=31$) & Open Targets ($n=34$) \\
    \midrule
    0 & $+0.0034$ & $+0.0133$ \\
    1 & $+0.0413$ & $+0.0542$ \\
    2 & $+0.0497$ & $+0.0719$ \\
    3 & $+0.0612$ & $+0.0658$ \\
    \bottomrule
    \end{tabular}
    \endgroup
\end{table}

\begin{figure}[!t]
    \centering
    \includegraphics[width=\columnwidth]{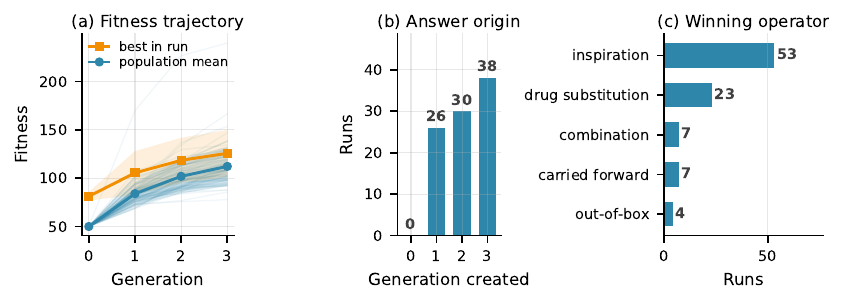}
    \caption{\textbf{Fitness trajectories and origins of final hypotheses.} Most final hypotheses are produced during search (87 of 94 through crossover or mutation), and fitness increases in every run. Panels show (a) mean and best fitness with shading for one standard deviation, (b) the generation of each final record, and (c) its generating operator; \emph{carried forward} denotes seven unchanged parent copies.}
    \label{fig:learning_curve}
\end{figure}

\noindent\textbf{Fitness improves in every run.} Fitness scores reflect the agents' pairwise judgments of hypotheses (Figure~\ref{fig:learning_curve}a). Population-mean fitness rises from 50.0 in generation 0 to 112.2 in generation 3, and best-member fitness rises from 81.2 to 125.7, with both increasing in all 94 runs. The initial population fixes the reference mean at 50 and the spread at 60 points. The largest gains occur in generations 1 and 2 under the agents' comparative judgments.

\noindent\textbf{Crossover and mutation produce most final hypotheses.} Figure~\ref{fig:methodology} illustrates one search trajectory, and Figure~\ref{fig:learning_curve}b,c summarizes the origins of the final hypotheses across runs. Of 94 final hypotheses, crossover produced 60, including 53 from inspiration; mutation produced 27, including 23 from drug substitution. The remaining 7 were unchanged parent copies created after an empty operator return. The final-output records were created in generation 1 for 26 runs, generation 2 for 30, and generation 3 for 38, including the seven unchanged copies.

\noindent\textbf{Search uses all four variation operators.} Across all 1{,}692 offspring records, inspiration accounts for 27.6\%, out-of-box mutation for 24.8\%, combination for 22.9\%, and drug substitution for 17.1\%. The remaining 7.5\% are unchanged parent copies after empty operator returns.

\noindent\textbf{Evolution refines rationales and changes drug choices.} In pancreatic adenocarcinoma, the leading hypothesis retains \drugname{olaparib} while narrowing a generic DNA-damage rationale to stratification by \emph{BRCA1}, \emph{BRCA2}, and \emph{PALB2}. In melanoma, the leading candidate changes from \drugname{trametinib} to \drugname{vemurafenib}. Both patterns occur across all three seeds. The first makes the conditions for a proposed intervention more specific; the second changes the intervention under consideration. These are generated hypotheses requiring experimental validation. The case studies in Appendix~\ref{app:qualitative} show how crossover combines parent explanations and mutation proposes an alternative mechanism.

\subsection{Ablations and Computational Cost}
\label{sec:ablations}

\begin{figure}[!t]
    \centering
    \includegraphics[width=\columnwidth]{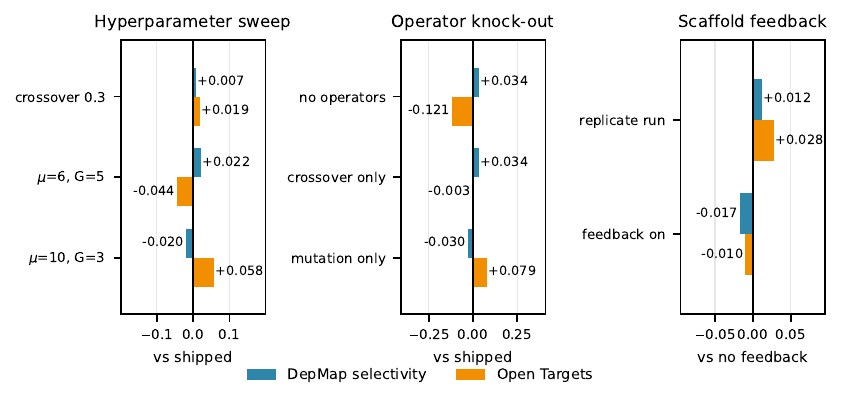}
    \caption{\textbf{Search settings, variation operators, and scaffold feedback.} Configuration changes (left) and operator removal (center) produce mixed shifts across the two metrics on eight cancer types; no comparison survives multiple-testing correction. For the co-scientist scaffold, feedback changes scores by amounts comparable to repeating the run (right). These scaffold comparisons cover 31 cancer types on \depmap and 34 on Open Targets.}
    \label{fig:ablations}
\end{figure}

\noindent\textbf{Larger populations and longer searches give mixed results.} Across eight cancer types, we test larger populations, longer horizons, and an alternative operator setting (Figure~\ref{fig:ablations}, left). Raising the population to $\mu=10$ changes \depmap selectivity by $-0.020$ and Open Targets association by $+0.058$. Extending the horizon to $G=5$ gives $+0.022$ and $-0.044$, while an operator setting with crossover probability 0.3 and mutation probability 0.4 gives $+0.007$ and $+0.019$. No setting differs significantly from the predefined configuration on either metric.

\noindent\textbf{Removing crossover or mutation has mixed effects.} We disable crossover, mutation, and both together on the same eight cancer types (Figure~\ref{fig:ablations}, center). Removing both reduces each generation to parent cloning and reranking, changing \depmap selectivity by $+0.034$ and Open Targets association by $-0.121$. The three conditions yield six comparisons across the two metrics, none of which survives Holm correction for multiple testing across the eight cancer types.

\noindent\textbf{Feedback has small effects in the co-scientist baseline.} We compare scaffold versions that provide or withhold feedback from the generator (Figure~\ref{fig:ablations}, right). Across 31 cancer types on \depmap selectivity and 34 on Open Targets, feedback changes the scores by $-0.017$ and $-0.010$, respectively. Repeating the withheld-feedback condition changes them by $+0.012$ and $+0.028$. The scaffold and static reranking also achieve similar scores in the main comparison (Figure~\ref{fig:depmap_summary}).

\noindent\textbf{Fitness-guided selection raises average and minimum scores.} We replace fitness-based parent selection with uniform random selection, holding generation, crossover, mutation, retrieval, evaluation, and the hypothesis count fixed. Across 31 cancer types on \depmap selectivity and 34 on Open Targets, fitness-guided selection raises the weakest member's scores by 0.088 and 0.218, and the population mean by 0.075 and 0.128. Maximum external scores show no statistically detectable change. For the final hypotheses, the margins are $+0.090$ on \depmap selectivity and $+0.156$ on Open Targets. The clearest effect is stronger biological support for the average and weakest hypotheses. Appendix~\ref{app:statistics} provides selection, configuration, and scaffold statistics.

\begin{table}[!t]
    \caption{\textbf{Computational cost of hypothesis discovery.} \oursystem uses fewer model calls than static reranking and the co-scientist scaffold, with pairwise scoring accounting for 167 of 206 calls. \emph{Model calls/run} and \emph{Cost/run} exclude retrieval; costs use run logs for \oursystem and single-pass generation and estimates from model-call counts for other methods.}
    \label{tab:cost}
    \centering
    \begin{tabular}{lrrr}
    \toprule
    Method & Candidates/run & Model calls/run & Cost/run \\
    \midrule
    Co-scientist scaffold & 15.0 & 288 & \$0.71 \\
    Static reranking (15 candidates) & 15.0 & 227 & \$0.55 \\
    \textbf{\oursystem} & \textbf{24.0} & \textbf{206} & \textbf{\$0.56} \\
    Self-consistency (40 samples) & 39.1 & 42 & \$0.21 \\
    Tree of Thoughts & 9.0 & 31 & \$0.09 \\
    Multi-agent debate & N/A & 17 & \$0.08 \\
    Single-pass generation & 1.0 & 1 & \$0.004 \\
    \bottomrule
    \end{tabular}
\end{table}

\phantomsection\label{sec:cost}
\noindent\textbf{More model calls do not consistently improve results.} \oursystem uses 206 model calls per run, including 167 pairwise comparisons (Table~\ref{tab:cost}). The co-scientist scaffold uses 288 calls and the static reranking control uses 227, reflecting the cost of comparing every pair in a 15-candidate pool. Both score within 0.007 of single-pass generation on \depmap selectivity. Tree of Thoughts offers the strongest lower-cost baseline, reaching 0.115 with 31 calls. With 40 samples, self-consistency scores below single-pass generation on both metrics. Costs use 94 \oursystem run logs and 204 single-pass run logs; other estimates use observed per-call prices. Token budgets \mbox{are not matched across methods}.

\section{Conclusion}

\oursystem formulates hypothesis development as a population-search problem for scientific agents. We couple reasoning over scientific claims and rationales with a generational genetic algorithm that directs selection, variation, and replacement. This formulation gives each agent contribution a defined role in the search and makes the rules of collaboration available for controlled study. Our drug repurposing evaluation connects the generated explanations to target-level biological evidence from two complementary sources. Across 34 cancer types, \oursystem achieves the highest mean scores among six baselines on both \depmap selectivity and Open Targets association. On the shared 26-type \depmap panel, selectivity is 0.171, compared with 0.039 for single-pass generation. Crossover or mutation produces 87 of the 94 final hypotheses.

The parent-selection study directly tests how a coordination decision affects hypothesis quality. Fitness-guided selection improves the population's mean and minimum scores on both external measures with scientific operations and hypothesis count held fixed. The result establishes a contribution of search design to the biological support for the hypotheses agents develop. It motivates a broader research agenda in which the organization of scientific teams is designed and evaluated alongside the capabilities of individual agents. By making collaboration itself a subject of algorithmic design, \oursystem creates a foundation for autonomous research teams with scientific capabilities \mbox{beyond those of individual models}.

\clearpage
\appendix
\section{\texorpdfstring{Discussion}{Discussion}}

\subsection{\texorpdfstring{Implications and Scope}{Implications and Scope}}

\oursystem separates reasoning about scientific claims from decisions about which hypotheses develop across generations. LLM agents supply scientific judgments and hypothesis transformations; the genetic algorithm coordinates their contributions through population updates. Because these updates are explicit, researchers can compare parent-selection and replacement rules with the scientific roles of the agents held fixed.

The drug repurposing study instantiates this principle with domain-specific retrieval, fitness criteria, semantic operators, and external evidence. The reusable component is the population-search architecture that coordinates those elements. Scientific agents already operate over single-cell data, cell-type annotation, and materials simulation \citep{gao2025scpilot,wang2026cellmaster,hu2026tritondft}. These domains illustrate how the same genetic control flow could be paired with task-specific representations and evidence sources. Evaluating this transfer across tasks and model families is an important direction for future work.

\subsection{\texorpdfstring{Limitations and Future Directions}{Limitations and Future Directions}}

The main practical constraint is computational cost. Pairwise comparison accounts for 167 of the 206 model calls in a run and grows quadratically with the evaluated pool. Sparse comparison schedules can draw on work in efficient comparative assessment \citep{liusie2024pairwise,yin2026decentralized} and redirect this budget toward larger populations or longer horizons. Fitness also depends on LLM judgment, so future versions can incorporate expert preferences or domain evidence directly into the comparison process. Agentic sequential falsification offers a route from free-form hypotheses to external tests \citep{huang2025popper}.

The current operator study covers eight cancer types, and broader replicated analyses can clarify the individual contributions of crossover and mutation. Adaptive operator rates and diversity-aware replacement may also help allocate search effort across hypothesis transformations \citep{lehman2011novelty,bradley2024qdaif}.

\subsection{\texorpdfstring{Impact Statement}{Impact Statement}}

\oursystem could help researchers compare competing explanations and identify drug repurposing hypotheses for experimental follow-up. The accompanying rationales and parent-offspring records make it possible to inspect which claims were retained, revised, or combined. LLM-based selection can also propagate unsupported assumptions across generations, directing attention toward persuasive hypotheses with weak evidence. Expert review of the supporting literature and independent experiments remain necessary before these proposals inform therapeutic decisions.

\clearpage
\section{Algorithm}
\label{app:algorithm}

Algorithm~\ref{alg:hypoevolve} formalizes the population update in Equation~\ref{eq:population-update}. The generation agent supplies $P_0$, and $\textsc{Score}$ compares every unordered pair in its input pool before fitting Bradley-Terry fitness as in Equation~\ref{eq:comparative-fitness}. Each size-2 tournament samples distinct candidates uniformly and returns the higher-fitness candidate. Crossover applies with probability $p_c=0.6$ and mutation with probability $p_m=0.15$, with mutation forced when crossover is skipped. Both operators choose uniformly between their two variants; an empty return invokes the unchanged-parent fallback described in Section~\ref{sec:genetic-search}. Each offspring receives a separate record with parentage and operator provenance, making both semantic transformations and fallback copies traceable across generations.

\begin{algorithm}[ht]
\caption{\oursystem}
\label{alg:hypoevolve}
\begingroup
\begin{algorithmic}[1]
\REQUIRE Research goal $g$; population size $\mu$; offspring count $\lambda$; horizon $G$
\REQUIRE Crossover rate $p_c$; mutation rate $p_m$
\ENSURE Highest-fitness hypothesis $h^*$ from the final population
\STATE $P_0 \leftarrow \textsc{Generate}(g,\mu)$
\STATE $f_0 \leftarrow \textsc{Score}(P_0)$ \COMMENT{Initialize the fitness scale}
\FOR{$t=1$ to $G$}
    \STATE $O_t \leftarrow \emptyset$
    \WHILE{$|O_t|<\lambda$}
        \STATE $p_1 \leftarrow \textsc{Tournament}(P_{t-1},f_{t-1},2)$
        \STATE $h_{\mathrm{new}} \leftarrow p_1$; $c \leftarrow \textsc{false}$
        \IF{$\operatorname{rand}()<p_c$}
            \REPEAT
                \STATE $p_2 \leftarrow \textsc{Tournament}(P_{t-1},f_{t-1},2)$
            \UNTIL{$p_2\ne p_1$}
            \STATE $h_{\mathrm{new}} \leftarrow \textsc{Crossover}(p_1,p_2)$; $c \leftarrow \textsc{true}$
        \ENDIF
        \IF{$c=\textsc{false}$ or $\operatorname{rand}()<p_m$}
            \STATE $h_{\mathrm{new}} \leftarrow \textsc{Mutation}(h_{\mathrm{new}})$
        \ENDIF
        \STATE $O_t \leftarrow O_t \uplus \{h_{\mathrm{new}}\}$ \COMMENT{Record offspring identity, parents, and operators}
    \ENDWHILE
    \STATE $Q_t \leftarrow P_{t-1}\uplus O_t$
    \STATE $f_t \leftarrow \textsc{Score}(Q_t)$ \COMMENT{Compare parents and offspring together}
    \STATE $P_t \leftarrow \operatorname{Top}_{\mu}(Q_t;f_t)$
\ENDFOR
\STATE \textbf{return} $\arg\max_{h\in P_G}f_G(h)$
\end{algorithmic}
\endgroup
\end{algorithm}

\FloatBarrier
\section{Additional Results}
\label{app:results}

\subsection{\texorpdfstring{Statistical Details for Main Results}{Statistical Details for Main Results}}
\label{app:statistics}

Table~\ref{tab:paired-details} details five baseline comparisons in Figure~\ref{fig:depmap_summary} and the held-out analysis in Section~\ref{sec:results}. Tables~\ref{tab:selection-details} and~\ref{tab:configuration-details} provide the selection-ablation statistics and configuration scores underlying Section~\ref{sec:ablations}. Each analysis uses the cancer panel stated in its table.

Evidence-channel tests use the 34-cancer panel in Section~\ref{sec:results}. Comparisons with single-pass generation give $p=1.4\times10^{-5}$ for known drugs and clinical trials, $p=1.3\times10^{-8}$ for literature, and $p=9.8\times10^{-6}$ for genetic association.

\clearpage
\begin{table}[!ht]
    \caption{\textbf{Statistical details for main and held-out comparisons.} \depmap comparisons favor \oursystem; the Open Targets difference from Tree of Thoughts remains statistically unresolved. \emph{Margin} reports mean score differences in favor of \oursystem, \emph{W/L} counts cancer-type wins and losses excluding ties, and $p$ denotes paired Wilcoxon tests.}
    \label{tab:paired-details}
    \centering
    \begingroup\color{black}
    \begin{tabular}{@{}lrrcrrc@{}}
    \toprule
    & \multicolumn{3}{c}{\depmap selectivity ($n=26$)} & \multicolumn{3}{c}{Open Targets ($n=29$)} \\
    Baseline & Margin & W/L & $p$ & Margin & W/L & $p$ \\
    \midrule
    Single-pass & $+0.1325$ & 19/7 & $3.2\times10^{-4}$ & $+0.2633$ & 26/3 & $1.1\times10^{-6}$ \\
    Self-consistency & $+0.1420$ & 18/7 & $0.0027$ & $+0.2926$ & 23/5 & $4.6\times10^{-5}$ \\
    Static reranking & $+0.1356$ & 17/8 & $0.0030$ & $+0.2513$ & 23/4 & $9.9\times10^{-5}$ \\
    Multi-agent debate & $+0.1015$ & 16/8 & $0.0056$ & $+0.2337$ & 24/3 & $8.1\times10^{-5}$ \\
    Tree of Thoughts & $+0.0567$ & 15/7 & $0.0113$ & $+0.0974$ & 14/10 & $0.126$ \\
    \midrule
    & \multicolumn{3}{c}{Held-out \depmap ($n=24$)} & \multicolumn{3}{c}{Held-out Open Targets ($n=27$)} \\
    Single-pass & $+0.1114$ & 18/6 & $2.8\times10^{-4}$ & $+0.2798$ & 25/2 & $8.2\times10^{-7}$ \\
    \bottomrule
    \end{tabular}
    \endgroup
\end{table}

\begin{table}[!ht]
    \caption{\textbf{Fitness-guided versus uniform parent selection.} Fitness-guided selection raises mean and minimum population scores on both external measures; population maxima show no statistically detectable change. Positive \emph{Margin} values favor fitness-guided selection with all other search components and hypothesis count fixed; \emph{Final hypothesis} reports the answer selected by the method.}
    \label{tab:selection-details}
    \centering
    \begingroup\color{black}
    \begin{tabular}{@{}lrrrr@{}}
    \toprule
    & \multicolumn{2}{c}{\depmap selectivity ($n=31$)} & \multicolumn{2}{c}{Open Targets ($n=34$)} \\
    Statistic & Margin & $p$ & Margin & $p$ \\
    \midrule
    Population minimum & $+0.0879$ & $6.6\times10^{-4}$ & $+0.2181$ & $1.2\times10^{-6}$ \\
    Population mean & $+0.0753$ & $8.3\times10^{-4}$ & $+0.1283$ & $1.0\times10^{-6}$ \\
    Population maximum & $+0.0154$ & $0.987$ & $+0.0017$ & $0.927$ \\
    Final hypothesis & $+0.0904$ & $0.064$ & $+0.1555$ & $0.0017$ \\
    \bottomrule
    \end{tabular}
    \endgroup
\end{table}

The configuration study uses eight cancer types selected before the runs, with one seed per type and setting. Four or five types retain the same drug across settings; increasing the population yields four wins and no losses on Open Targets.

\begin{table}[!ht]
    \caption{\textbf{Search configuration sensitivity on eight cancer types.} No setting differs significantly from the default on either metric. Unspecified parameters retain $\mu=6$, $G=3$, $p_c=0.6$, and $p_m=0.15$; $p$ values compare each setting with the default.}
    \label{tab:configuration-details}
    \centering
    \begingroup\color{black}
    \begin{tabular}{@{}lrrrr@{}}
    \toprule
    & \multicolumn{2}{c}{\depmap selectivity} & \multicolumn{2}{c}{Open Targets} \\
    Configuration & Mean & $p$ & Mean & $p$ \\
    \midrule
    Default & $0.0939$ & N/A & $0.4495$ & N/A \\
    Population $\mu=10$ & $0.0743$ & $0.144$ & $0.5079$ & $0.068$ \\
    Horizon $G=5$ & $0.1157$ & $0.715$ & $0.4052$ & $0.273$ \\
    $p_c=0.3$, $p_m=0.4$ & $0.1013$ & $0.593$ & $0.4685$ & $0.285$ \\
    \bottomrule
    \end{tabular}
    \endgroup
\end{table}

The scaffold feedback test uses three rounds of five candidates and one seed per cancer type. Compared with withholding feedback, providing it yields 3 wins and 10 losses on \depmap ($n=31$, $p=0.173$), and 6 wins and 8 losses on Open Targets ($n=34$, $p=0.778$). Approximately 60\% of cancer types receive the same drug under both conditions.

\clearpage
\subsection{\texorpdfstring{Fitness Improves Across All Cancer Types}{Fitness Improves Across All Cancer Types}}

\begin{figure}[!ht]
    \centering
    \includegraphics[width=\columnwidth]{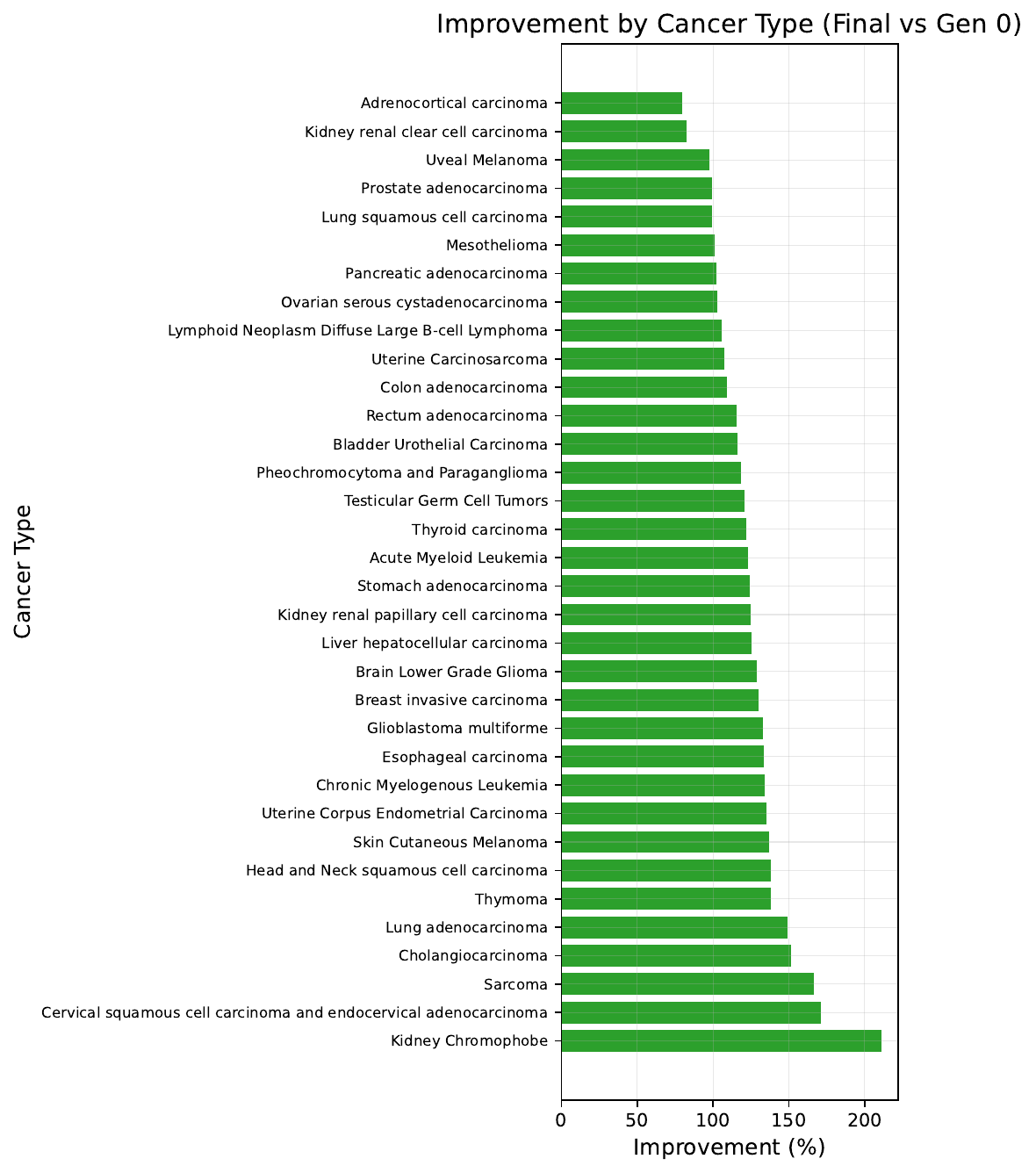}
    \caption{\textbf{Fitness improvement by cancer type.} Mean fitness increases in all 34 cancer types, showing consistent progress under the agents' comparative assessments. Bars show percentage changes from generation 0 to generation 3 relative to the common initial fitness scale.}
    \label{fig:app_improvement}
\end{figure}

Fitness improves for every evaluated cancer type (Figure~\ref{fig:app_improvement}), consistent with the aggregate trajectory in Figure~\ref{fig:learning_curve}. Generation 0 is scaled the same way in every run, with the population mean centred at 50 and the spread set to 60 points, so the percentages below are measured against a common baseline. Mean fitness rises in 94 of 94 runs, by 124.2\% on average ($p = 1.7 \times 10^{-49}$, paired $t$-test), and best fitness rises in 94 of 94 runs, by 54.9\% ($p = 1.3 \times 10^{-32}$).

\FloatBarrier
\clearpage
\subsection{\texorpdfstring{\depmap Selectivity Improves in Most Cancer Types}{DepMap Selectivity Improves in Most Cancer Types}}

\begin{figure}[!ht]
    \centering
    \includegraphics[width=\columnwidth]{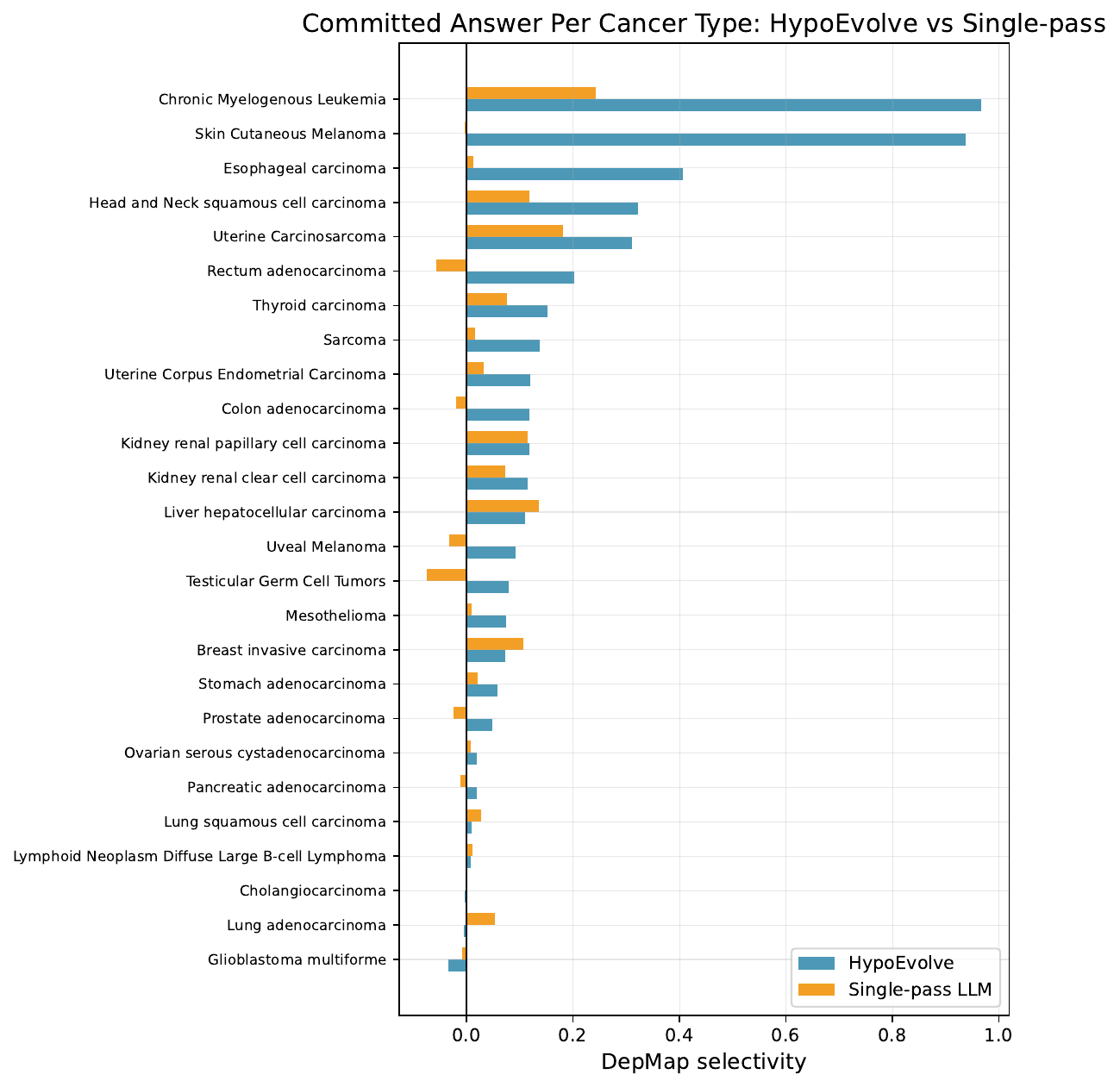}
    \caption{\textbf{\depmap selectivity by cancer type.} \oursystem exceeds single-pass generation in 19 of 26 cancer types, showing that the aggregate advantage extends to most evaluated contexts. Bars compare per-cancer scores for \oursystem (blue) and single-pass generation (orange). Positive selectivity indicates greater target dependency in the matched cancer than in the pan-cancer reference.}
    \label{fig:app_depmap}
\end{figure}

\oursystem exceeds single-pass generation on \depmap selectivity in 19 of the 26 cancer types shared by all methods and \depmap; single-pass generation leads in 7 (Figure~\ref{fig:app_depmap}). Scores average the hypotheses selected by each method across its runs for each cancer type, following Section~\ref{sec:setup}. This comparison shows how the aggregate advantage in Figure~\ref{fig:depmap_summary} varies across cancer contexts.

\FloatBarrier
\clearpage
\subsection{\texorpdfstring{Evolution Improves Drug-Cancer Matching}{Evolution Improves Drug-Cancer Matching}}
\label{app:specificity}

We examine whether hypothesis development improves the match between a drug and a cancer context after accounting for the drug's overall score. For an external metric $s$, let $\mathcal{C}_s$ be the cancer types with available scores. The cancer-specificity residual for a proposed drug $d$ and cancer $c$ is
\begin{equation}
r_s(d,c)=s(d,c)-\frac{1}{|\mathcal{C}_s|-1}\sum_{\substack{c'\in\mathcal{C}_s\\c'\ne c}}s(d,c').
\label{eq:specificity}
\end{equation}
This adjustment removes a drug's average advantage across other cancer contexts. The analysis covers 31 types on \depmap selectivity and 34 on Open Targets, using the available \oursystem results. The main comparison uses the smaller common set covered by all methods.

From generation 0 to generation 3, the residual increases in 21 of 31 cancer types on \depmap selectivity ($p=0.0071$) and 25 of 34 on Open Targets ($p=0.0017$). The final-generation means are $+0.0612$ and $+0.0658$, respectively (Table~\ref{tab:specificity}). These results support improvement within \oursystem. Comparisons with four multi-candidate baselines favor \oursystem directionally on both residuals, but none survives correction for multiple testing.

A constant-drug control illustrates why cancer specificity requires separate assessment. We select one drug using the seven development cancer types, freeze that choice, and evaluate it on the 27 held-out types. Its Open Targets mean is 0.4827, compared with 0.4290 for \oursystem, with 8 wins and 17 losses for \oursystem ($p=0.045$). \depmap selectivity favors \oursystem by $+0.0917$, with 14 wins and 8 losses ($p=0.017$). On Open Targets, the constant drug exceeds Tree of Thoughts by 0.1735 and self-consistency by 0.3596.

\subsection{\texorpdfstring{Comparison and Scoring Details}{Comparison and Scoring Details}}
\label{app:comparison-details}

The task-matched comparison shares the backbone, drug vocabulary, three literature-search queries, and output format. Each method commits to its answer before external scoring. For single-pass generation, scores average six independent draws within each cancer type.

The main Tree of Thoughts configuration uses the shared retrieval protocol. A second configuration without retrieval obtains mean \depmap selectivity of 0.1183 and Open Targets association of 0.3556. Paired comparisons with \oursystem on this variant's evaluated cancer set give margins of $+0.0577$ on \depmap ($p=0.0061$) and $+0.0651$ on Open Targets ($p=0.489$). Tree of Thoughts remains the strongest alternative, and the Open Targets margin for this configuration is not statistically resolved.

A separate diagnostic selects the externally best of the six single-pass draws. The mean margins in favor of \oursystem are $+0.032$ on \depmap selectivity ($p=0.87$) and $+0.120$ on Open Targets ($p=0.019$). This comparison assesses the sampled pool using external evidence unavailable to the methods during answer selection; the \depmap difference remains statistically unresolved.

Drugs map to curated target genes, and cancer types map to \depmap cell lines through \texttt{OncotreePrimaryDisease}. The LGG/GBM, COAD/READ, and KIRC/KIRP pairs resolve to identical cell-line sets. \depmap selectivity is computed within the resulting matched sets, with each target's pan-cancer median subtracted before taking the maximum over targets. These mappings specify each comparison's biological coverage and target-level interpretation. Under this selectivity metric, drug-by-cancer interaction accounts for 79.34\% of score variance, and ten distinct drugs attain the maximum across cancer contexts.

\FloatBarrier
\clearpage
\subsection{\texorpdfstring{Qualitative Examples of Hypothesis Development}{Qualitative Examples of Hypothesis Development}}
\label{app:qualitative}

\begingroup
\tcbset{hypothesisexample/.style={
    colback=white,colbacktitle=black!4,colframe=black!30,coltitle=black,
    boxrule=0.45pt,arc=1.2mm,boxsep=0pt,
    left=9pt,right=9pt,top=8pt,bottom=8pt,
    toptitle=5pt,bottomtitle=5pt,
    fonttitle=\bfseries,fontupper=\small,
    before skip=0pt,after skip=10pt
}}

\begin{figure}[!ht]
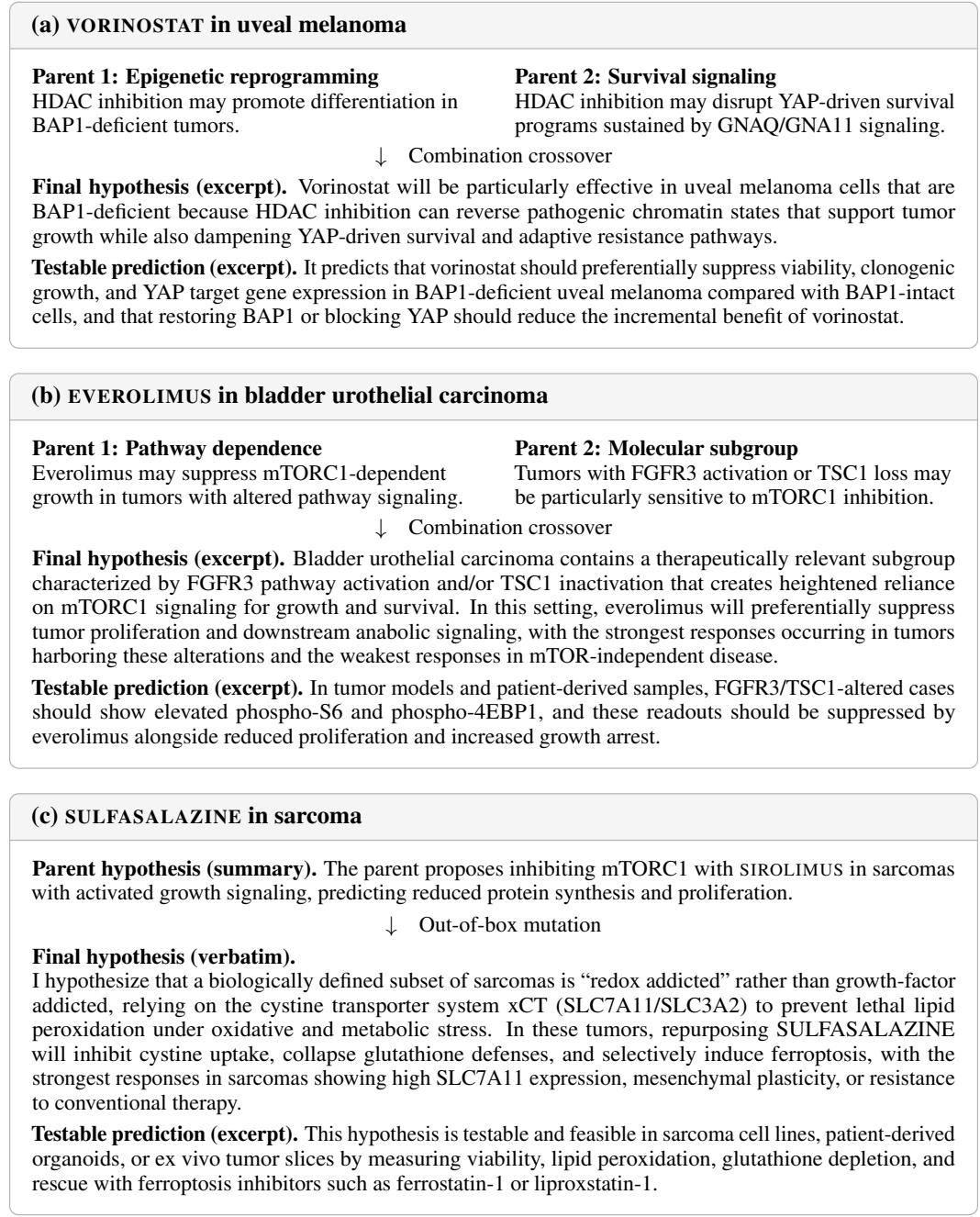

\centering
\begin{tcolorbox}[hypothesisexample,title={(a) \drugname{Vorinostat} in uveal melanoma}]
\begin{minipage}[t]{0.477\linewidth}
\raggedright
\textbf{Parent 1: Epigenetic reprogramming}\par
 HDAC inhibition may promote differentiation in BAP1-deficient tumors.
\end{minipage}\hfill
\begin{minipage}[t]{0.477\linewidth}
\raggedright
\textbf{Parent 2: Survival signaling}\par
 HDAC inhibition may disrupt YAP-driven survival programs sustained by GNAQ/GNA11 signaling.
\end{minipage}

\smallskip
{\centering$\downarrow$\quad Combination crossover\par}
\smallskip
\textbf{Final hypothesis (excerpt).} Vorinostat will be particularly effective in uveal melanoma cells that are BAP1-deficient because HDAC inhibition can reverse pathogenic chromatin states that support tumor growth while also dampening YAP-driven survival and adaptive resistance pathways.

\smallskip
\textbf{Testable prediction (excerpt).} It predicts that vorinostat should preferentially suppress viability, clonogenic growth, and YAP target gene expression in BAP1-deficient uveal melanoma compared with BAP1-intact cells, and that restoring BAP1 or blocking YAP should reduce the incremental benefit of vorinostat.

\end{tcolorbox}

\begin{tcolorbox}[hypothesisexample,title={(b) \drugname{Everolimus} in bladder urothelial carcinoma}]
\begin{minipage}[t]{0.477\linewidth}
\raggedright
\textbf{Parent 1: Pathway dependence}\par
 Everolimus may suppress mTORC1-dependent growth in tumors with altered pathway signaling.
\end{minipage}\hfill
\begin{minipage}[t]{0.477\linewidth}
\raggedright
\textbf{Parent 2: Molecular subgroup}\par
 Tumors with FGFR3 activation or TSC1 loss may be particularly sensitive to mTORC1 inhibition.
\end{minipage}

\smallskip
{\centering$\downarrow$\quad Combination crossover\par}
\smallskip
\textbf{Final hypothesis (excerpt).} Bladder urothelial carcinoma contains a therapeutically relevant subgroup characterized by FGFR3 pathway activation and/or TSC1 inactivation that creates heightened reliance on mTORC1 signaling for growth and survival. In this setting, everolimus will preferentially suppress tumor proliferation and downstream anabolic signaling, with the strongest responses occurring in tumors harboring these alterations and the weakest responses in mTOR-independent disease.

\smallskip
\textbf{Testable prediction (excerpt).} In tumor models and patient-derived samples, FGFR3/TSC1-altered cases should show elevated phospho-S6 and phospho-4EBP1, and these readouts should be suppressed by everolimus alongside reduced proliferation and increased growth arrest.

\end{tcolorbox}

\begin{tcolorbox}[hypothesisexample,title={(c) \drugname{Sulfasalazine} in sarcoma}]
\textbf{Parent hypothesis (summary).} The parent proposes inhibiting mTORC1 with \drugname{sirolimus} in sarcomas with activated growth signaling, predicting reduced protein synthesis and proliferation.

\smallskip
{\centering$\downarrow$\quad Out-of-box mutation\par}
\smallskip
\textbf{Final hypothesis (verbatim).}\par
 I hypothesize that a biologically defined subset of sarcomas is \mbox{``redox addicted''} rather than growth-factor addicted, relying on the cystine transporter system xCT (SLC7A11/SLC3A2) to prevent lethal lipid peroxidation under oxidative and metabolic stress. In these tumors, repurposing SULFASALAZINE will inhibit cystine uptake, collapse glutathione defenses, and selectively induce ferroptosis, with the strongest responses in sarcomas showing high SLC7A11 expression, mesenchymal plasticity, or resistance to conventional therapy.

\smallskip
\textbf{Testable prediction (excerpt).} This hypothesis is testable and feasible in sarcoma cell lines, patient-derived organoids, or ex vivo tumor slices by measuring viability, lipid peroxidation, glutathione depletion, and rescue with ferroptosis inhibitors such as ferrostatin-1 or liproxstatin-1.
\end{tcolorbox}

\caption{\textbf{Qualitative examples of hypothesis development.} Crossover connects epigenetic and survival-signaling arguments in (a) and links pathway dependence to molecular conditions for drug response in (b). Mutation shifts the proposed vulnerability from growth signaling to antioxidant defense in (c), with a mechanistic rescue test. Parent arguments are summarized; final hypotheses and testable predictions are reproduced as verbatim excerpts.}
\label{fig:qualitative}
\end{figure}

\noindent\textbf{Literature context.} Published experiments in osteosarcoma report that \drugname{sulfasalazine} lowers glutathione, increases lipid peroxidation, and induces cell death that can be rescued by ferroptosis inhibitors \citep{guo2023mlx}. These findings support the proposed mechanism in osteosarcoma; extending the prediction to other sarcoma subtypes requires direct testing.
\endgroup

\FloatBarrier
\clearpage
\section{Prompt Templates}
\label{app:prompts}

This section presents the core prompts used by \oursystem agents for the drug repurposing task.

\subsection{Generation Agent Prompt}

The generation prompt supplies the research goal, supporting literature, and required output format.

\begin{tcolorbox}[colback=gray!5,colframe=gray!65,colbacktitle=gray!65,coltitle=white,title=Generation Prompt,fonttitle=\bfseries\small,fontupper=\small]
You are an expert tasked with formulating a novel and robust hypothesis to address the following objective. You have conducted a thorough review of relevant literature and developed a logical framework for addressing the objective.

\textbf{Goal:} \{goal\}

\textbf{Criteria for a strong hypothesis:} \{preferences\}

\textbf{Literature review and analytical rationale:} \{articles\_with\_reasoning\}

\textbf{Required Output Format:}

\texttt{TITLE:} [A concise, descriptive title]

\texttt{SUMMARY:} [Single-sentence summary]

\texttt{HYPOTHESIS:} [Clear statement in 2-3 sentences]

\texttt{RATIONALE:} [Detailed explanation including key mechanisms, evidence from literature, and testability]

\texttt{FINAL DRUG:} [Drug name]

\texttt{CANCER TYPE:} [TCGA cancer type]
\end{tcolorbox}

\subsection{Pairwise Comparison Prompt}

The comparison prompt ranks two hypotheses for the same cancer and allows a tie. Its judgments determine the fitness used for parent selection and population replacement.

\begin{tcolorbox}[colback=gray!5,colframe=gray!65,colbacktitle=gray!65,coltitle=white,title=Pairwise Comparison Prompt,fonttitle=\bfseries\small,fontupper=\small]
Compare two drug repurposing hypotheses for the SAME cancer type and pick the stronger one.

A stronger hypothesis is one whose proposed drug acts on a dependency that is SPECIFIC to this cancer type, a lineage-defining oncogene, a mutated or amplified driver, or a pathway this tumour type is selectively addicted to.

\textbf{Judge on:}
\begin{enumerate}[leftmargin=*,itemsep=0pt,topsep=0pt]
\item \textbf{Specificity}. Would this drug plausibly work better in THIS cancer than in an arbitrary other cancer? A mechanism that applies equally to every tumour type is WEAKER, not stronger, because it does not explain why this cancer was chosen.
\item \textbf{Target evidence}. Is the named target actually implicated in this cancer type?
\item \textbf{Testability}. Does the hypothesis make a concrete, falsifiable prediction?
\end{enumerate}

Explicitly DO NOT reward: generic cytotoxicity, broadly pleiotropic agents, or mechanisms that reduce to ``this pathway matters in cancer generally''.

\textbf{Respond in exactly this format:}

\texttt{REASONING:} <one to three sentences explaining your choice>

\texttt{CONFIDENCE:} <HIGH | MEDIUM | LOW>

\texttt{WINNER=}<A | B | TIE>
\end{tcolorbox}

\clearpage
\subsection{Evolution Agent Prompts}

\noindent\textbf{Combination Crossover.} The agent integrates scientific content from selected parent hypotheses.

\begin{tcolorbox}[colback=gray!5,colframe=gray!65,colbacktitle=gray!65,coltitle=white,title=Combination Crossover Prompt,fonttitle=\bfseries\small,fontupper=\small]
You are synthesizing a unified hypothesis from multiple parent hypotheses.

\textbf{Goal:} \{goal\}

\textbf{Parent hypotheses:} \{hypotheses\}

\textbf{Review feedback:} \{reviews\}

\textbf{Instructions:} Integrate the strongest aspects from each parent into a coherent unified hypothesis. Preserve beneficial mechanisms while addressing identified weaknesses. The offspring should be superior to any individual parent.
\end{tcolorbox}

\noindent\textbf{Out-of-Box Mutation.} The agent reconsiders assumptions to develop an alternative explanation.

\begin{tcolorbox}[colback=gray!5,colframe=gray!65,colbacktitle=gray!65,coltitle=white,title=Out-of-Box Mutation Prompt,fonttitle=\bfseries\small,fontupper=\small]
You are generating a novel hypothesis inspired by but distinct from provided concepts.

\textbf{Goal:} \{goal\}

\textbf{Inspiration (use analogy, not replication):} \{hypotheses\}

\textbf{Instructions:}
\begin{enumerate}[leftmargin=*,itemsep=0pt,topsep=0pt]
\item Identify promising avenues for exploration
\item Develop a detailed, original hypothesis leveraging analogous principles
\item This should not be a mere aggregation of existing methods; think out-of-the-box
\end{enumerate}
\end{tcolorbox}

\subsection{Drug Constraint}

Each drug repurposing hypothesis must select from 61 FDA-approved drugs with known targets covered by \depmap CRISPR data, enabling consistent external assessment across methods.

\begin{tcolorbox}[colback=gray!5,colframe=gray!65,colbacktitle=gray!65,coltitle=white,title=Drug Constraint,fonttitle=\bfseries\small,fontupper=\small]
You MUST select your drug repurposing candidate ONLY from this approved list:

\drugname{SIMVASTATIN}, \drugname{ATORVASTATIN}, \drugname{METFORMIN}, \drugname{HYDROXYCHLOROQUINE}, \drugname{PROPRANOLOL}, \drugname{SERTRALINE}, \drugname{OMEPRAZOLE}, \drugname{ASPIRIN}, \drugname{CELECOXIB}, \drugname{DOXYCYCLINE}, \drugname{DISULFIRAM}, \drugname{THALIDOMIDE}, \drugname{SIROLIMUS}, \drugname{EVEROLIMUS}, \drugname{IMATINIB}, \drugname{DASATINIB}, \drugname{SORAFENIB}, \drugname{ERLOTINIB}, \drugname{VEMURAFENIB}, \drugname{OLAPARIB}, \drugname{VENETOCLAX}, \drugname{IBRUTINIB}, \drugname{PALBOCICLIB}, \drugname{RUXOLITINIB}, ... [61 drugs total]

These drugs have been verified to have: (1) FDA approval, (2) known target genes in Open Targets Platform, (3) target genes present in DepMap CRISPR data.
\end{tcolorbox}

\end{document}